# Edge detection based on morphological amoebas [1]

W. Y. Lee[a], Y. W. Kim[a], S. Y. Kim[b], J. Y. Lim[b], and D. H. Lim[c2]

[a] Pohang University of Science and Technology, Pohang 790-784, Korea
[b] Korea Advanced Institute of Science and Technology, Daejeon 305-701, Korea
[c] Department of Information Statistics and RINS, Gyeongsang National University, Jinju 660-701, Korea

### Abstract

Detecting the edges of objects within images is critical for quality image processing. We present an edge-detecting technique that uses morphological amoebas that adjust their shape based on variation in image contours. We evaluate the method both quantitatively and qualitatively for edge detection of images, and compare it to classic morphological methods. Our amoeba-based edge-detection system performed better than the classic edge detectors.

**Keywords:** edge detection; morphological amoeba; structuring element

## 1 INTRODUCTION

An object edge is a boundary at which a significant change of gray level occurs in the image. Information related to the position, shape, size, and surface design of a subject in an image can be obtained using an edge-detection algorithm. Edges are generally detected using gradient techniques such as Sobel, Prewitt, Roberts, Laplacian of Gaussian, and Canny operators [1, 2], but these are not suitable for detecting edges in noisy images because both noise and edge have high optical frequency. The images used in many practical applications (e.g., medical imaging) are usually corrupted by noise, and thus separating edges from noise and other interference is difficult but necessary.

Statistical approaches, such as those proposed by Bovik et al. [3], Lim and Jang [4], and Lim [5], offer an alternative to gradient techniques. The concept behind these statistical approaches is to examine the distribution of intensity values in the neighborhood of a given pixel and determine if the pixel should be classified as an edge.

Mathematical morphology provides yet another approach to image processing, one based on shape concepts originating from set theory [6]. Images in mathematical morphology theory are treated as sets, and morphological transformations derived from Minkowski addition and subtraction are defined to extract features in images. Morphological edge detectors have been studied for use in applications where the performance of classic edge detectors is reduced because of noise [7]. For example, Lee et al. [8] proposed a blur-minimization (BM) edge detector, which minimizes erosion and dilation residues of blurred images. Feehs and Arce [9] showed the importance of blurring an original image for morphological edge detection. They introduced an $\alpha$-trimmed morphological (ATM) edge detector that incorporates opening and closing operations. They also proved statistically that the ATM edge detector performs better than the BM edge detector. Zhao et al. [10] proposed the reduced noise morphological (RNM) edge detector for edge detection in medical images. They demonstrated that it was more efficient for de-noising and detecting edges than the more commonly used classic edge detectors such as the Sobel and Laplacian of Gaussian operators, and general morphological edge detectors such as the morphological gradient (MG) edge detector and dilation residue edge detector [11].

---

[1] This research was supported by Basic Science Research Program through the National Research Foundation of Korea(NRF) funded by the Ministry of Education, Science and Technology(No. 2010-0016815)

[2] Corresponding author : Department of Information Statistics and RINS, Gyeongsang National University, Jinju 660-701, Korea; dhlim@gnu.ac.kr

Mathematical morphology develops elements with certain structures and features for measuring and processing images. Conventional morphological edge detectors perform operations by applying a fixed, space-invariant structuring element (SE) to all image pixels, although the size and shape of the SE may have been selected arbitrarily. This can be problematic because the local properties of the input pixels may not be identical throughout the image and thus the fixed SE may not accurately process all areas of the image. The selection and optimization of the SE is a difficult challenge, and is a hot topic in the field of mathematical morphology research. *Does a SE that changes its shape according to image features exist?* If so, it would not matter what kind of SEs were used for edge detection and one would not have to take the local properties of input image pixels into account. The answer to this question is, yes, such a SE does exist: the *amoeba*. In fact, the term amoeba was first used by Lerallut et al. [12] who proposed morphological operators with amoebas for *noise reduction*.

There have been several studies related to spatially-variant mathematical morphology, i.e. mathematical morphology with spatially-variant SEs. Theoretical basis of spatially-variant mathematical morphology was studied in [13], and several specific examples of this morphology were introduced in [12, 14–17]. General adaptive neighborhood morphology [14], proposed using the concept of criterion mapping and a tolerance, is one example of spatially-variant morphology. Mathematical morphology with morphological amoebas is an another example of spatially-variant morphology. Although spatially-variant morphology was applied to image filtering [12, 13–15], denoising [12, 13], segmentation [13, 14], enhancement [14], restoration [16], and feature extraction [17], there have been no works on edge detection with spatially-variant morphology.

Here, we describe a new edge-detecting technique that uses amoebas that adapt their shape based on variation in image contours. The proposed edge detector uses *modified* amoebas that are one pixel larger than the original ones used for noise reduction applications.

The remainder of this paper is organized as follows. In Section 2, we review classic SEs and edge detectors with classic SEs. In Section 3, we present original amoebas, which are dynamic SEs for noise reduction, and introduce modified amoebas for edge detection. Furthermore, we describe edge detection using modified amoebas. Section 4 presents a number of experimental results to demonstrate the performance of our edge detector using amoebas. We present our conclusions in Section 5.

## 2 CLASSIC MORPHOLOGICAL EDGE DETECTION

*2.1 Classic structuring elements*

Until now, the shapes and sizes of SEs have been determined by repeated trial-and-error experiments. That is, one can determine the appropriate SEs for the particular geometric shapes in an image by testing various elements with various shapes on many images. For example, circular SEs are commonly applied in biological or medical images without sharp angles and straight lines, while square SEs are often used in images with many straight lines, e.g., aerial photographs of cities. Large SEs preserve big features in images and small SEs preserve fine features. Hence, big SEs are normally used to reduce noise; if SEs are too small, noise will be recognized as image features. However, if SEs are too large, image restoration performance is degraded because image details are removed. As previously mentioned, there are no morphological algorithms with special SEs that perform well for all images; different SEs are required for images with different features. That is, an algorithm that has excellent performance for one image usually leads to poor results for another image.

*2.2 Edge detection with classic structuring elements*

In this subsection, we review four existing morphological edge detectors that use single symmetrical SEs. Before doing so, however, we require some preliminary definitions of morphological operations. The basic morphological operations are dilation, erosion, and combinations of these two operations. The dilation and erosion of a function $f$ by a spatially-invariant SE $B$ are defined by:

$$\text{dilation: } (f \oplus B)(x) = \max\{f(z) | z \in B(x)\},$$
$$\text{erosion: } (f \ominus B)(x) = \min\{f(z) | z \in B(x)\}.$$

Based on these, the opening $f \circ B$ and closing $f \bullet B$ of a function $f$ with respect to $B$ are defined by:

$$\text{opening: } f \circ B = (f \ominus B) \oplus B,$$
$$\text{closing: } f \bullet B = (f \oplus B) \ominus B.$$

*2.2.1 MG edge detector*

The MG edge detector, which was proposed by Beucher [11], is defined by

$$f_o(z) = (f \oplus B)(z) - (f \ominus B)(z), \tag{1}$$

where $(f \oplus B)(z)$ and $(f \ominus B)(z)$ are dilation and erosion, respectively, of the image $f(z)$ with SE $B$, and $f_o(z)$ is the output image.

*2.2.2 BM edge detector*

The BM edge detector, which was developed by Lee et al. [8], is defined by

$$f_o(z) = \min\{f_{av}(z) - (f_{av} \ominus B)(z), (f_{av} \oplus B)(z) - f_{av}(z)\}, \qquad (2)$$

where $f_{av}(z)$ is the input image blurred with a two-dimensional running mean filter, $f_o(z)$ is the output image, and $B$ is the SE, which is a square with sides of length $2n+1$.

*2.2.3 ATM edge detector*

Feehs and Arce [9] proposed the ATM edge detector, which replaces the mean filter of Eq. (2) with the $\alpha$-trimmed mean filter [18]. The ATM edge detector is defined as

$$f_o(z) = \min\{(f_\alpha \circ B)(z) - (f_\alpha \ominus B)(z), (f_\alpha \oplus B)(z) - (f_\alpha \bullet B)(z)\}, \qquad (3)$$

where $f_\alpha(z)$ is the input image blurred with a two-dimensional running $\alpha$-trimmed mean filter, $(f_\alpha \circ B)(z)$ and $(f_\alpha \bullet B)(z)$ are opening and closing, respectively, of the $f_\alpha(z)$ with SE $B$.

*2.2.4 RNM edge detector*

Zhao et al. [10] first used the opening-closing operation in preprocessing to remove noise as

$$M(z) = ((f \bullet B) \circ B)(z). \qquad (4)$$

They then smoothed the image by first closing and then dilating. The RNM edge detector is based on the difference between the processed image using this process and the image before dilation, and defined as

$$f_o(z) = ((M \bullet B) \oplus B - M \bullet B)(z). \qquad (5)$$

## 3 PROPOSED EDGE DETECTION USING AMOEBAS

*3.1 Amoebas for noise reduction*

An amoeba is a type of protozoan without definite form. It consists of a mass of protoplasm containing one or more nuclei surrounded by a flexible outer membrane. The term amoeba is sometimes used to refer to something with an indefinite, changeable shape.

As mentioned above, morphological amoebas for noise reduction was introduced in [12]. The amoeba was used as a dynamic SE that changes shape to adapt to variation in image contours. The shape of the amoeba must be computed for each pixel around which it is centered. Figure 1(b) shows the shape of an amoeba that depends on the position of its center. Note that in flat areas such as the center of the disc or the background, the amoeba is maximally stretched, and is reluctant to cross contour lines.

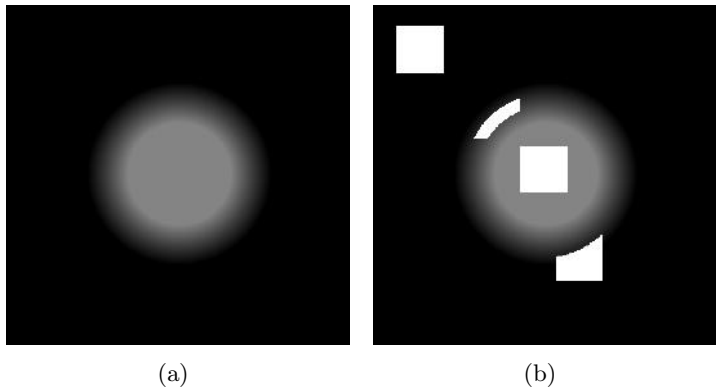

(a)                                 (b)

Figure 1: Shapes of amoebas at various positions on an image: (a) original image and (b) shapes of amoebas

The shape of the amoeba is calculated by the *amoeba distance*, which is defined as follows: Let $d_{pixel}(x, y)$ be a difference of intensities between pixel $x$ and $y$ and $\sigma = (x = x_0, x_1, \ldots, x_n = y)$ be a path between $x$ and $y$. Then, the length of path $\sigma$ is defined as

$$L_\lambda(\sigma) = \sum_{i=1}^{n}[1 + \lambda \cdot d_{pixel}(x_{i-1}, x_i)], \quad \text{where } \lambda \text{ is a real number.}$$

Thus, the amoeba distance $d_\lambda$ can be defined with parameter $\lambda$ where

$$\begin{cases} d_\lambda(x, x) = 0 \\ d_\lambda(x, y) = \min_\sigma L_\lambda(\sigma). \end{cases}$$

The amoeba distance is based on both the geometric and pixel value distance. The parameter $\lambda$ represents the degree of difference between two pixel values. We can thus define an amoeba at pixel $x$ in the following expression using the amoeba distance:

$$Amoeba_{\lambda,r}(x) = \{y | d_\lambda(x, y) \leq r\}, \tag{6}$$

where $r$ is a real number that denotes the amoeba radius. The amoeba is square if $\lambda = 0$.

*3.2 Amoebas for edge detection*

Despite the amoeba defined above was successfully applied for noise reduction in [12], there must be serious problems if we apply it for edge detection. Consider a simple image in Figure 2(a) with a marked center pixel. The amoeba centered at the marked pixel would spread its shape out as Figure 2(b); when $r = 3$ with normal $\lambda$ (at least 0.1), the shape of amoeba would be like Figure 2(c). It means that the amoeba cannot cross over an edge

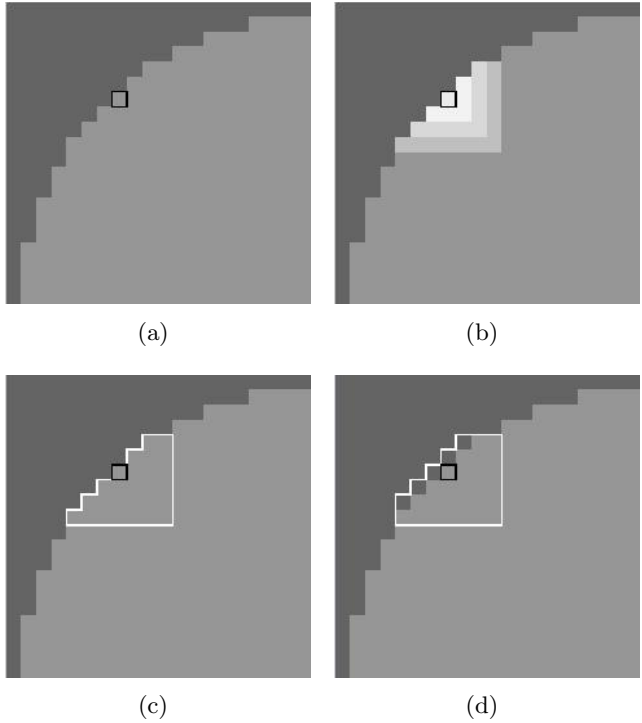

Figure 2: Shapes of amoebas for noise reduction and edge detection ($r = 3$, $\lambda \geq 0.1$): (a) original image, (b) spreading out amoeba, (c) an amoeba for noise reduction (d) a modified amoeba for edge detection

in this Figure 2(c). However, with this amoebas, which cannot cross edges, we cannot get 'any' edges in this image using even MG edge detector, the simplest morphological edge detector. The reason is that the difference between dilation and erosion at each pixel would be exactly zero because every amoeba includes only same-valued pixels. In other words, to detect edges using morphological methods, SEs should contain pixels over edges.

There must be another problem if we set large $r$ and small $\lambda$ in order to make amoebas cross over edges. In this case, to detect obvious edges like an edge between white and black region, we have to use much bigger $r$ and smaller $\lambda$. It causes amoebas to contain many dissimilar pixels such as noises; finally, these amoebas would have no difference between normal spatially-invariant SEs, and morphological edge detectors with these amoebas would be noise-sensitive.

To solve these problems, we propose a *modified* amoeba $Amoeba^+_{\lambda,r}(x)$, one pixel larger than the original one $Amoeba_{\lambda,r}(x)$ given in Eq. (6), as follows.

$$Amoeba^+_{\lambda,r}(x) = \{y \big| y \in Amoeba_{\lambda,r}(x) \oplus H \text{ and } |x - y| \leq r\}, \qquad (7)$$

where $H$ is a 3×3 diamond-shaped structuring element of 0's. The modified amoeba is enforced to be expanded one pixel from the original one in order to cross over edges. As a result, the amoeba with appropriate $r$ and $\lambda$ could detect edges well while it mostly contains similar pixels. Figure 2(d) shows the modified amoeba centered at the marked pixel. In the next subsection, edge detectors will be proposed using the modified amoeba rather than the original one.

*3.3 Edge detection with amoebas*

Dilation, erosion, opening, and closing using our amoebas with some modification will be referred to as *amoeba dilation*, *amoeba erosion*, *amoeba opening*, and *amoeba closing*, respectively. The amoeba dilation and amoeba erosion of a function $f$ by the modified amoeba $A_{\lambda,r}^+$ are defined respectively by

$$\text{amoeba dilation: } (f \oplus_\beta A_{\lambda,r}^+)(x) = k\text{-th max}\{f(z)|z \in Amoeba_{\lambda,r}^+(x)\},$$
$$\text{amoeba erosion: } (f \ominus_\beta A_{\lambda,r}^+)(x) = k\text{-th min}\{f(z)|z \in Amoeba_{\lambda,r}^+(x)\},$$

where $Amoeba_{\lambda,r}^+(x)$ denotes an modified amoeba at pixel $x$ in Eq. (7), $\beta$ is a positive constant, and $k = \lceil \beta \cdot |Amoeba_{\lambda,r}^+(x)| \rceil$. The expression $\lceil x \rceil$ denotes the smallest integer not less than $x$, and $|x|$ in $\lceil \ \rceil$ denotes the number of pixels in $x$. Note that the amoeba dilation and amoeba erosion are defined by $k$-th min and $k$-th max rather than min and max. The reason is that during performing dilation on the original amoeba to get the modified amoeba, some noisy pixels could be included in the modified amoeba, and the number of noisy pixels in the modified amoeba would be proportional to its size; thus, we have to exclude those noisy pixels by taking $k$-th min and $k$-th max when calculating amoeba dilation and amoeba erosion. Based on these, the amoeba opening and amoeba closing of a function $f$ by the modified amoeba $A_{\lambda,r}^+$ are also defined respectively by

$$\text{amoeba opening: } f \circ_\beta A_{\lambda,r}^+ = (f \ominus_\beta A_{\lambda,r}^+) \oplus_\beta A_{\lambda,r}^+,$$
$$\text{amoeba closing: } f \bullet_\beta A_{\lambda,r}^+ = (f \oplus_\beta A_{\lambda,r}^+) \ominus_\beta A_{\lambda,r}^+.$$

Here we describe our proposed edge detectors by applying these amoeba operations to the edge detectors discussed in Section 2.2. The amoeba-based edge detectors will use the *pilot image*, on which the shape of amoebas are computed, as in [12]. After computing the shape of amoebas on the *pilot image*, we perform amoeba operations on the *original* image.

*3.3.1 Amoeba MG edge detector*

The amoeba MG edge detector corresponding to Eq. (1) is defined as

$$f_o(z) = (f \oplus_\beta A_{\lambda,r}^+)(z) - (f \ominus_\beta A_{\lambda,r}^+)(z),$$

where the pilot image is a Gaussian blurred image of $f$.

*3.3.2 Amoeba BM edge detector*

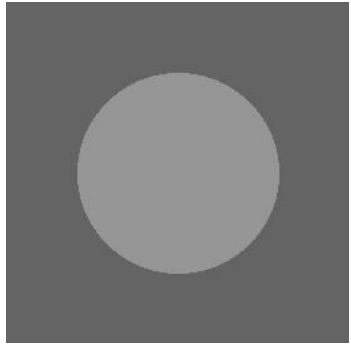

Figure 3: Artificial image used for testing and comparison purposes.

The amoeba BM edge detector corresponding to Eq. (2) is defined as

$$f_o(z) = \min\{f_{av}(z) - (f_{av} \ominus_\beta A^+_{\lambda,r})(z), (f_{av} \oplus_\beta A^+_{\lambda,r})(z) - f_{av}(z)\},$$

where the pilot image is a Gaussian blurred image of $f_{av}$.

*3.3.3 Amoeba ATM edge detector*

The amoeba ATM edge detector corresponding to Eq. (3) is defined as

$$f_o(z) = \min\{(f_\alpha \circ_\beta A^+_{\lambda,r})(z) - (f_\alpha \ominus_\beta A^+_{\lambda,r})(z), (f_\alpha \oplus_\beta A^+_{\lambda,r})(z) - (f_\alpha \bullet_\beta A^+_{\lambda,r})(z)\},$$

where the pilot image is a Gaussian blurred image of $f_\alpha$.

*3.3.4 Amoeba RNM edge detector*

The amoeba RNM edge detector corresponding to Eq. (5) is defined as

$$f_o(z) = ((M^+ \bullet_{\beta_2} A^+_{\lambda,r}) \oplus_{\beta_2} A^+_{\lambda,r} - M^+ \bullet_{\beta_2} A^+_{\lambda,r})(z),$$

where $M^+(z) = ((f \bullet_{\beta_1} A^+_{\lambda,r}) \circ_{\beta_1} A^+_{\lambda,r})(z)$ corresponding to Eq. (4), which represents the filtered image from the amoeba-based opening-closing operation. In this case, the pilot image for each amoeba operation is a Gaussian blurred image of the image to which each operation is applied.

## 4 EXPERIMENTAL RESULTS AND DISCUSSION

In this section, we compare amoeba-based morphological edge detection with classic morphological edge detection and the well-known Canny edge detector[2] when used on both artificial and real images. We use our amoebas with $\lambda = 1/2$ and $\beta = 0.1$ for MG, BM and ATM edge detectors, and $\beta_1 = 0.3$, $\beta_2 = 0.1$ for RNM edge detector. Several edge detection experiments were conducted on corrupted images using the two noise types, impulse noise and Gaussian noise [4, 5]. We conducted following experiments on a laptop with 2.40GHz Intel Core i5 and 4GB memory.

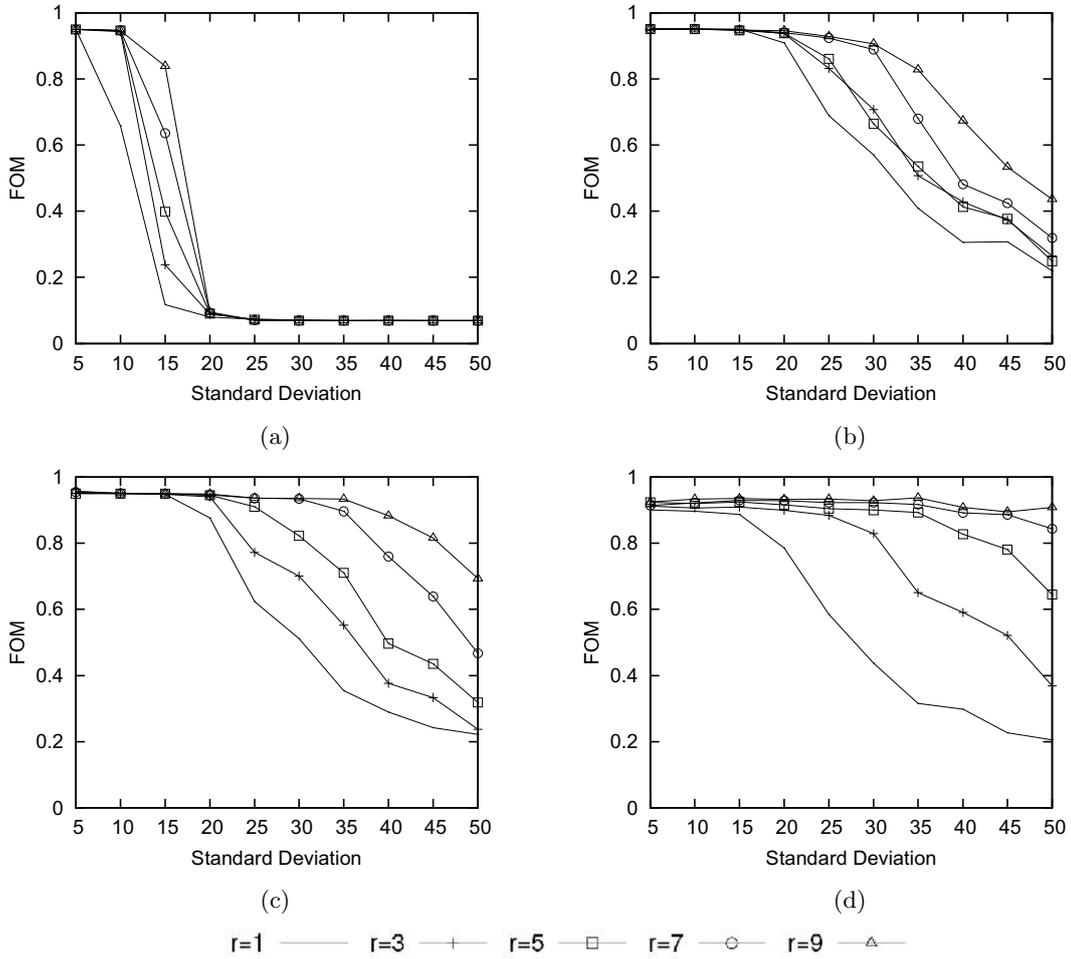

Figure 4: FOM results of amoeba-based edge detectors with several $r$ values in images with Gaussian noise: (a) Amoeba MG, (b) Amoeba BM, (c) Amoeba ATM, (d) Amoeba RNM.

The amoeba-based edge detectors were evaluated both quantitatively and qualitatively.

*4.1 Quantitative evaluation*

Pratt's figure of merit (FOM) [19] and the received operating characteristic (ROC) curve [20, 21] were used as performance measures for quantitative evaluation and comparison to the other edge detectors. The FOM is defined as

$$\text{FOM} = \frac{1}{\max\{I_I, I_D\}} \sum_{i=1}^{I_D} \frac{1}{1 + \alpha\,(d_i)^2},$$

where $I_I$ and $I_D$ are the number of ideal and detected edge points, respectively, and $d_i$

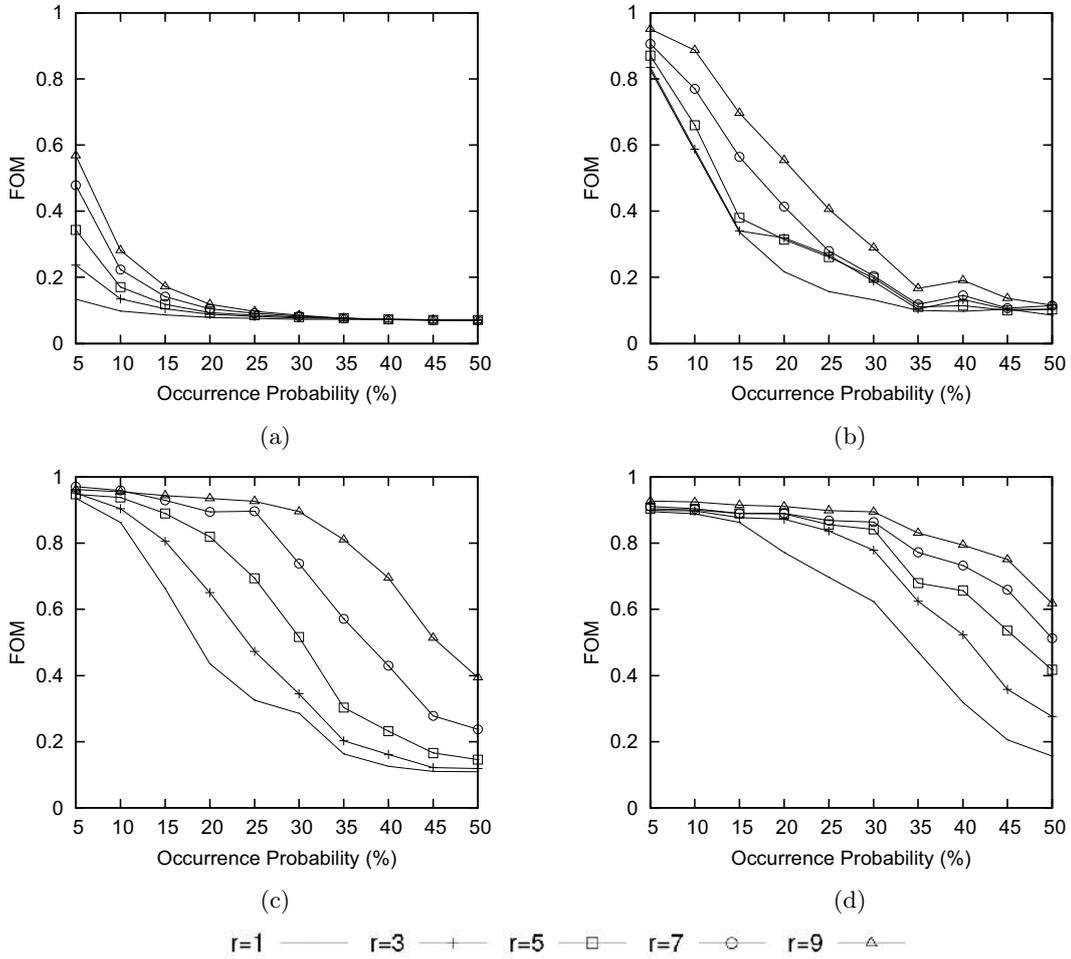

Figure 5: FOM results of amoeba-based edge detectors with several $r$ values in images with impulse noise: (a) Amoeba MG, (b) Amoeba BM, (c) Amoeba ATM, (d) Amoeba RNM.

is the distance between $i$th detected edge point and an ideal edge. The scaling constant $\alpha(>0)$ provides a relative tradeoff among smearing, isolation, and the edge offset, and was set to $\alpha = 1/9$. A FOM = 1 corresponds to a perfect match between the ideal and detected edge point, and the FOM approaches zero as the deviation of the detected points increases.

A ROC curve is the plot of the correct detection probability $P_d$ against the false detection probability $P_f$ for the different possible thresholds of an edge detector. More specifically, if $G$ represents the set of ideal edge points and $D$ represents the set of de-

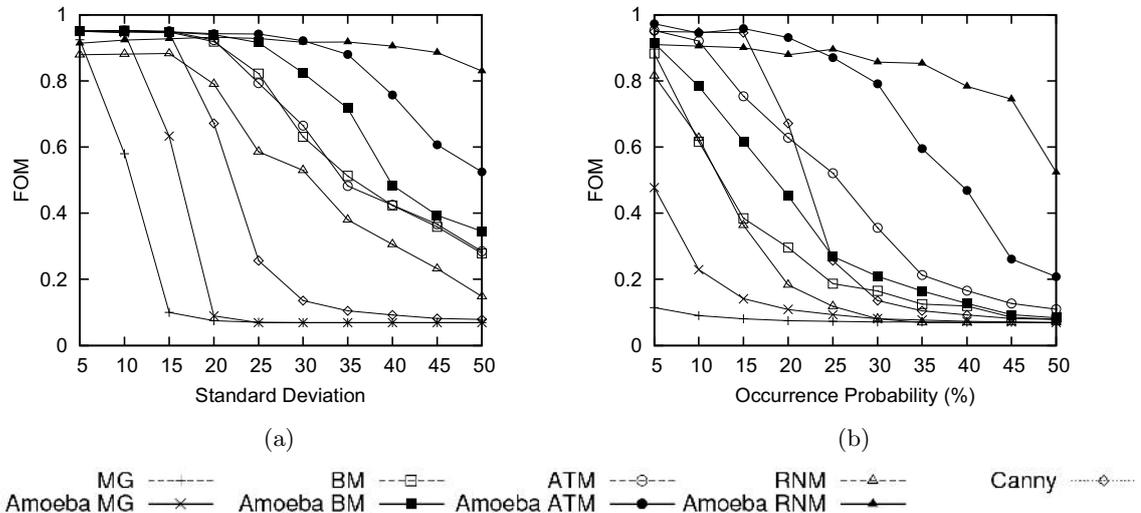

Figure 6: Measured values of FOM for (a) Gaussian noises, (b) Impulse noises.

tected edge points, then $P_d$ and $P_f$ is given by

$$P_d = \frac{n(D \cap G)}{n(G)}, \; P_f = \frac{n(D \cap G^c)}{n(G^c)},$$

where $n$ represents the number of pixels of the corresponding set. The area under the ROC curve can be used as an index to measure the performance of edge detectors, where a larger area under the curve represents better detector performance.

An artificial image was created and used as a benchmark to assess the performance of our amoeba-based edge-detection technique and compare it to other methods. We used a 256×256 image containing a circle, in which the outer panel had a gray level of 100 and the inner panel had a gray level of 150, as shown in Figure 3.

Figure 4 and 5 show the FOM results of amoeba-based edge detectors with respect to $r$ values in noisy images with Gaussian noise and impulse noise, respectively. In both figures, FOM value increases as $r$ value increases for all amoeba edge detectors. FOM values of amoeba MG edge detector is dropped rapidly as noise ratio increases, for the original MG edge detector itself is extremely noise-sensitive. For Gaussian noise, amoeba BM edge detector with $r \geq 7$ keeps its maximum performance when $\sigma \leq 25$; amoeba ATM and RNM edge detectors with $r \geq 7$ also keeps maximum FOM values when $\sigma \leq 30$ and $\sigma \leq 35$, respectively. For impulse noise, amoeba ATM and RNM edge detectors with $r \geq 7$ nearly maintain its maximum FOM values when $p \leq 25\%$ and $p \leq 30\%$, respectively. Although the performance of amoeba-based edge detectors is improved as $r$ grows, large $r$ value makes these methods have heavy computational time and ignore the details of an image. Therefore, we will use $r = 7$ in the following experiments.

Figure 6 shows the measured values of FOM used to compare amoeba edge detection to classic edge detection and Canny edge detector for Gaussian noise and impulse noise, respectively. All amoeba-based edge detection have better performance than classic edge detection for both Gaussian and impulse noise with entire noise ranges, $5 \leq \sigma \leq 50$ and $5\% \leq p \leq 50\%$. Moreover, as noise ratio increases, the performance of Canny edge detector drops sharply, meaning that it is not much robust. The amoeba MG edge detector is more robust than the classic MG edge detector, although the performance of both MG detectors declines significantly as noise ratio increases since MG detector is highly sensitive to noise. It is noteworthy that amoeba RNM detector is especially well performing for both noises than the original one; amoeba RNM detector even has the best FOM values in most case while RNM detector completely does not. Note that the ATM edge detector performs better than the BM detector, as mentioned in Feehs and Arce [9].

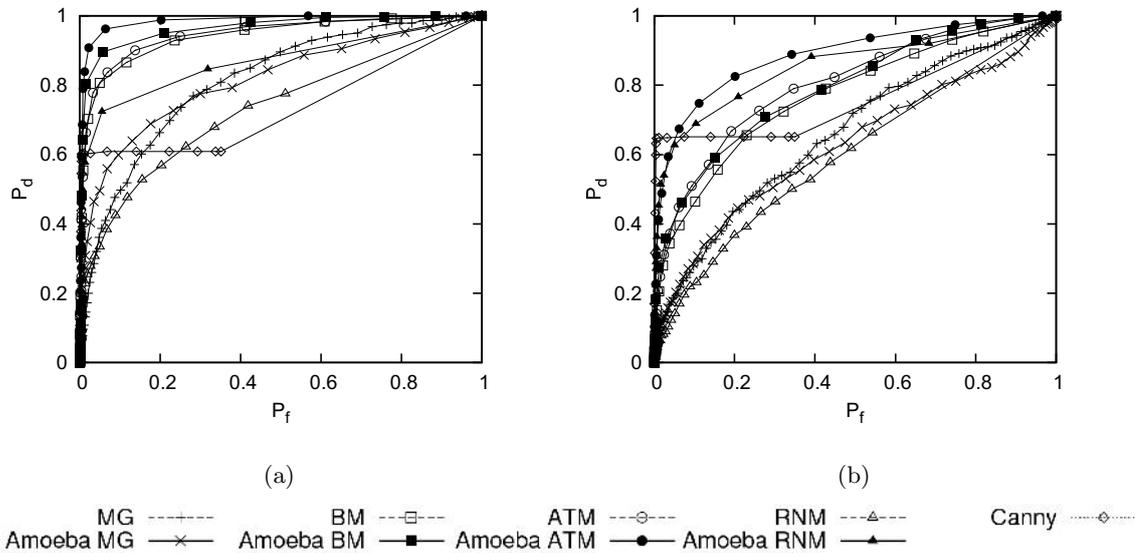

Figure 7: ROC curves for Gaussian noise with (a) $\sigma = 25$, (b) $\sigma = 45$.

Figure 7 compares the ROC curves for amoeba-based edge detection, classic edge detection and Canny edge detector for $\sigma = 25$ and $\sigma = 45$. Amoeba MG detector performs as well as MG detector, and amoeba BM detector shows slightly better performance than BM detector. Moreover, in ROC curves, not only amoeba RNM detector but also amoeba ATM detector, the best one in ROC curves, have significantly improved performance compared to corresponding classic detectors and Canny detector. Canny edge detector shows almost constant $P_d$ even when $P_f$ varies, which is somewhat different pattern from other edge detectors. This is probably due to the postprocessing stages of non-maxima suppression and hysteresis thresholding.

*4.2 Qualitative evaluation*

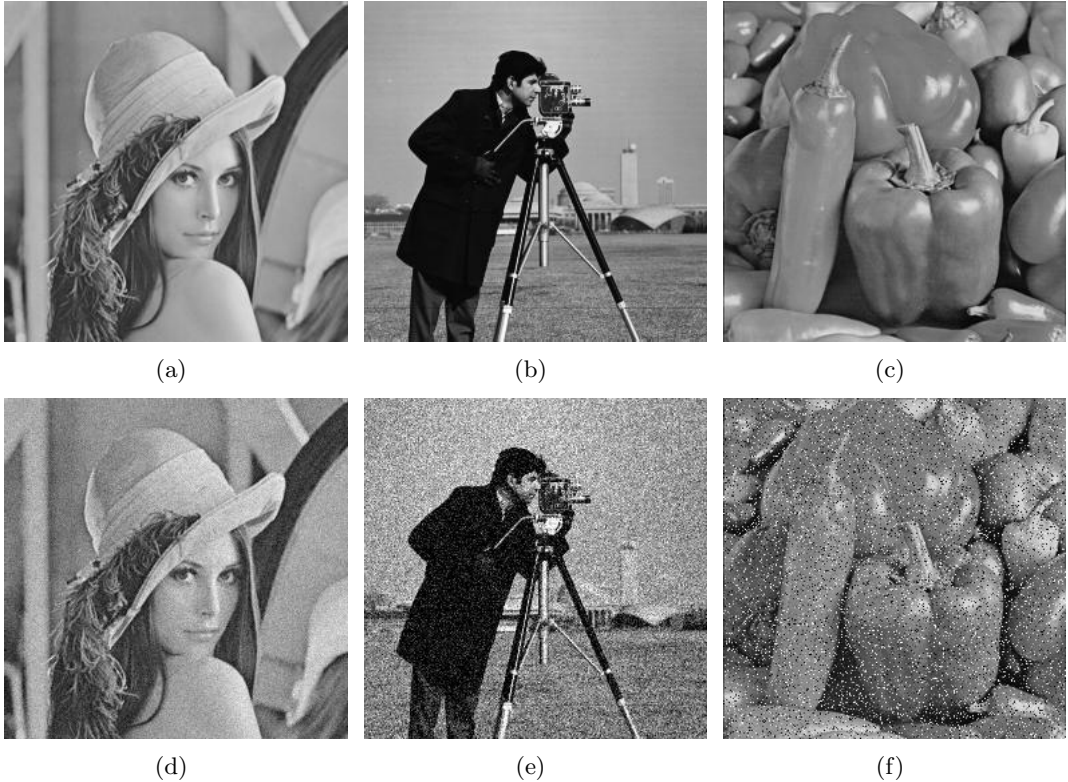

Figure 8: Images used in experiments: (a) Lena, (b) cameraman, (c) peppers, (d) Lena image corrupted with Gaussian noise with $\sigma = 10$, (e) cameraman image corrupted with Gaussian noise with $\sigma = 30$, and (f) peppers image corrupted with 20% impulse noise

We used the real 256×256 Lena, cameraman, and pepper images shown in Figure 8(a)–8(c), and noisy images corrupted with various degrees of Gaussian and impulse noise. Figure 8(d)–8(f) show some of the noisy images used in this experiment.

Figure 10 compares the results of edge detection using fixed SEs and amoebas, and Canny edge detector for the Lena image corrupted with $\sigma = 10$ Gaussian noise. The edge maps of the MG and amoeba MG edge detectors, illustrated in Figure 10(a) and 10(b), show that amoeba MG detector is more insensitive to noise than MG detector which made considerable noise speckling on the whole edge map. The difference in performance between BM and amoeba BM edge detector is difficult to distinguish in the maps, as shown in Figure 10(c) and 10(d). However, Figure 10(e) and 10(f) show that amoeba ATM edge detector detects more details, such as the left pillar and the background curve on the rightmost side, than ATM detector does. Similarly, Figure 10(g) and 10(h) also show that amoeba RNM detector is more sensitive to details of an image than RNM detector; only amoeba RNM detector detects edges of the left pillar, the pillar above the hat, and the background curve on the rightmost side. On the other hand, Canny edge detector catches more details compared to other edge detectors as shown in Figure 10(i).

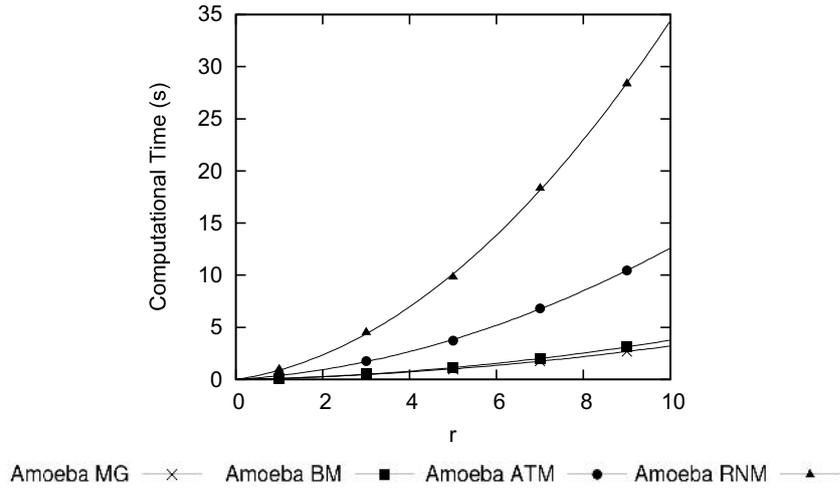

Figure 9: Computational times (in seconds) of amoeba edge detectors for the peppers image in Figure 8(f).

Figure 11 shows the results of edge detection based on fixed SEs and amoebas, and Canny edge detector for the cameraman image corrupted with $\sigma = 30$ Gaussian noise. Neither the MG nor the amoeba MG edge detectors could distinguish any meaningful edges due to the large amount of noise (see Figure 11(a) and 11(b)). However, the amoeba BM, ATM, and RNM edge detectors suppress more noise while preserving more details of the edges than their corresponding classic edge detectors (see Figure 11(c)–11(h)). Furthermore, amoeba ATM edge detector detects the handle of the camera and the building on the right side of a background more clearly than classic ATM detector does. Unlike Figure 10(i), Figure 11(i) is highly corrupted by noise, indicating that Canny edge detector is sensitive to noise.

Figure 12 shows the results of edge detection based on fixed SEs and amoebas, and Canny edge detector for the pepper image corrupted with 20% impulse noise. The results of MG, amoeba MG and Canny edge detector, Figure 12(a), 12(b) and 12(i), are highly contaminated with noise like Figure 11(a), 11(b) and 11(i). Figure 12(c)–12(h), however, demonstrate that amoeba BM, ATM and, especially, RNM are more robust to impulse noise than classic edge detectors. In addition, some edges, such as the bright regions of peppers, are detected more clearly by amoeba ATM detector than classic ATM detector.

4.3 Computational time

Figure 9 shows the computational time of amoeba edge detectors to process the corrupted peppers image in Figure 8(f), when $r$ varies. The curves in Figure 9 is the results

– 14 –

Table 1. Computational times (in seconds) of various edge detectors

| Edge Detectors | Test images | | |
|:---:|:---:|:---:|:---:|
| | Figure 8(d) | Figure 8(e) | Figure 8(f) |
| MG | 0.014 | 0.015 | 0.014 |
| Amoeba MG | 2.190 | 1.828 | 1.747 |
| BM | 0.016 | 0.017 | 0.016 |
| Amoeba BM | 2.458 | 2.126 | 2.007 |
| ATM | 0.072 | 0.073 | 0.071 |
| Amoeba ATM | 7.203 | 6.259 | 6.613 |
| RNM | 0.058 | 0.062 | 0.060 |
| Amoeba RNM | 19.47 | 19.12 | 18.07 |
| Canny | 0.037 | 0.057 | 0.037 |

of polynomial fitting with order 2 with respect to $r$. Since the shape of amoeba can be computed by a Dijkstra's region growing algorithm[22] using a binary heap, the time complexity of amoeba edge detectors should be $O(NM \cdot r^2 \log r)$, where $N$ and $M$ are the pixel number of image's width and height, respectively. Figure 9 demonstrates that theoretic time complexity is pretty consistent with the experimental results.

Table 1 provides the computational times of amoeba edge detectors, classic edge detectors and Canny edge detector for the three test images in Figure 8(d)–8(f). While classic edge detectors and Canny detector run in less than 0.1s, amoeba edge detectors run in more than about 2s. Especially amoeba ATM and RNM have much longer computational time than amoeba MG and BM in that they use more amoeba operations. This computational heaviness of amoeba operations are consistent with the results in [12].

## 5 CONCLUSION AND FUTURE WORK

Edge detection is an important first step in many image-processing applications. As such, it is important that the information derived from the edge-detection process should be as accurate as possible to ensure that higher-level processing is not affected.

Conventional morphological edge detectors use a fixed SE on all image pixels, although the size and shape of the SE may be arbitrary. This presents problems because the local properties of the input pixels may not be identical throughout the image and thus the fixed SE may not be appropriate to accurately process all areas of the image.

We presented a morphological edge-detection technique using spatially variant SEs, or modified amoebas, which adapt their shape to variation in image contours. Although morphological amoeba was already researched in noise reduction area, we proposed the modified amoeba, one pixel larger than the original one, in that it is difficult to apply the original amoeba directly for edge detection. Moreover, we proposed morphological amoeba-based operations, which use modified amoebas as spatially variant SEs, by re-placing min and max with $k$-th min and $k$-th max, for some noise pixels could be included in modified amoeba during the expansion of the original amoeba. Several amoeba-based edge detectors were finally proposed by using the amoeba operations.

We compared the performance of our proposed methods quantitatively and qualitatively to classic morphological edge detectors and Canny edge detector on both artificial and real images. First, it was found that the performance of amoeba-based edge detectors increased as $r$ value increased; however, we used $r = 7$ in the whole experiments to preserve details and make good trade-off between performance and computation time. The experimental results showed that our amoeba-based edge detectors performed better in terms of Pratt's FOM and ROC than classic edge detectors for artificial images with either Gaussian or impulse noise. The edge maps produced from real images indicated that our edge detectors were more robust to noise while detecting more details of images clearly than classic edge detectors. Especially, the performance of amoeba ATM and RNM detectors was significantly improved both quantitatively and qualitatively, compared to the corresponding classic edge detectors.

We also compared the computational time of proposed methods and classic edge detectors. First, the computational time of proposed methods was approximately proportional to $O(r^2 \log r)$. Second, proposed methods were much computationally heavier than classic edge detectors while their edge detection performances were better.

We plan to conduct more research to reduce computational time of amoeba edge detectors by using other techniques, such as the minimum spanning tree of an image, as in [23].

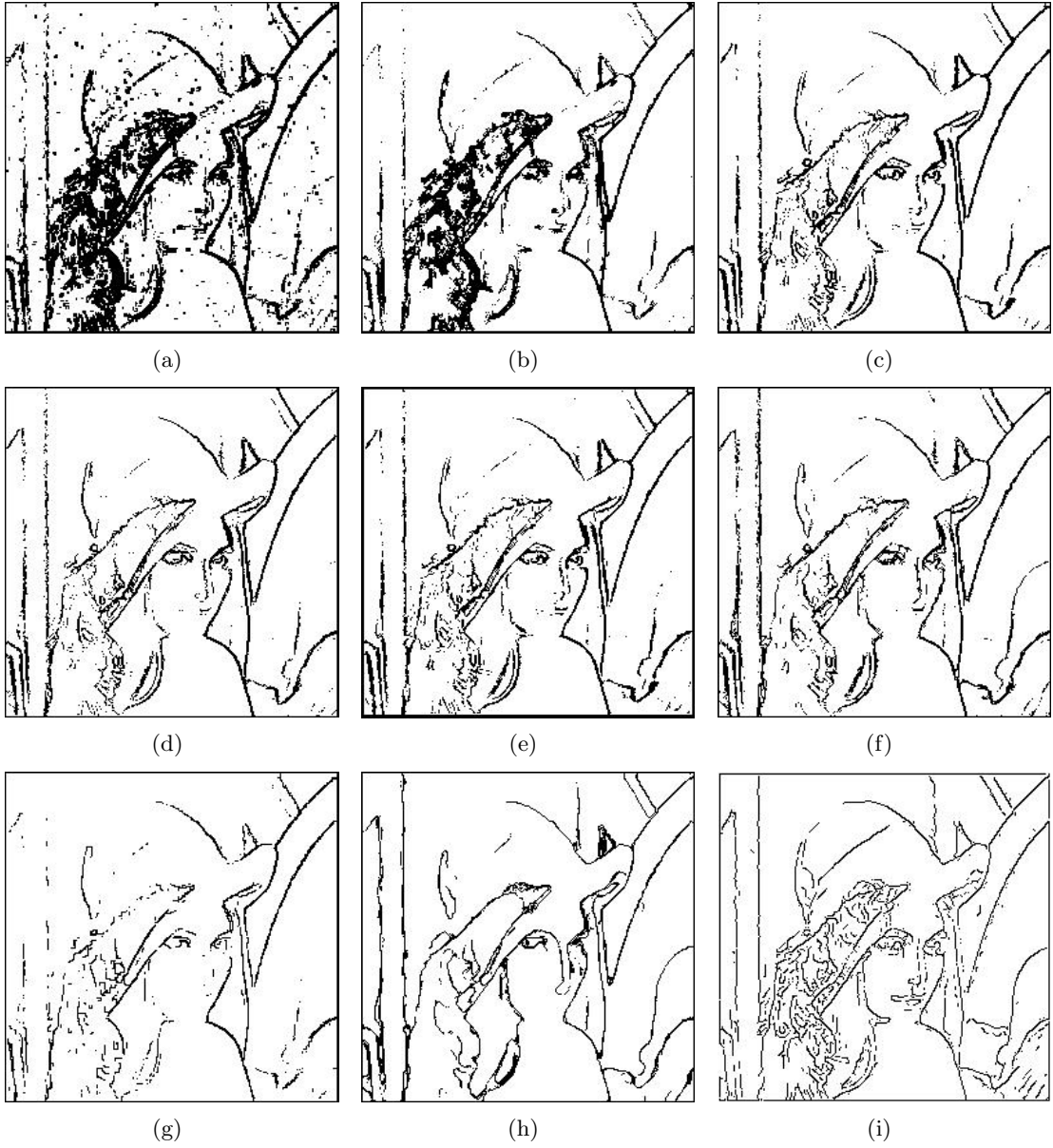

Figure 10: Edge detection result for Lena image in Figure 8(d): (a) MG detector, (b) Amoeba MG detector, (c) BM detector, (d) Amoeba BM Detector, (e) ATM detector, (f) Amoeba ATM detector, (g) RNM Detector, (h) Amoeba RNM detector, and (i) Canny detector

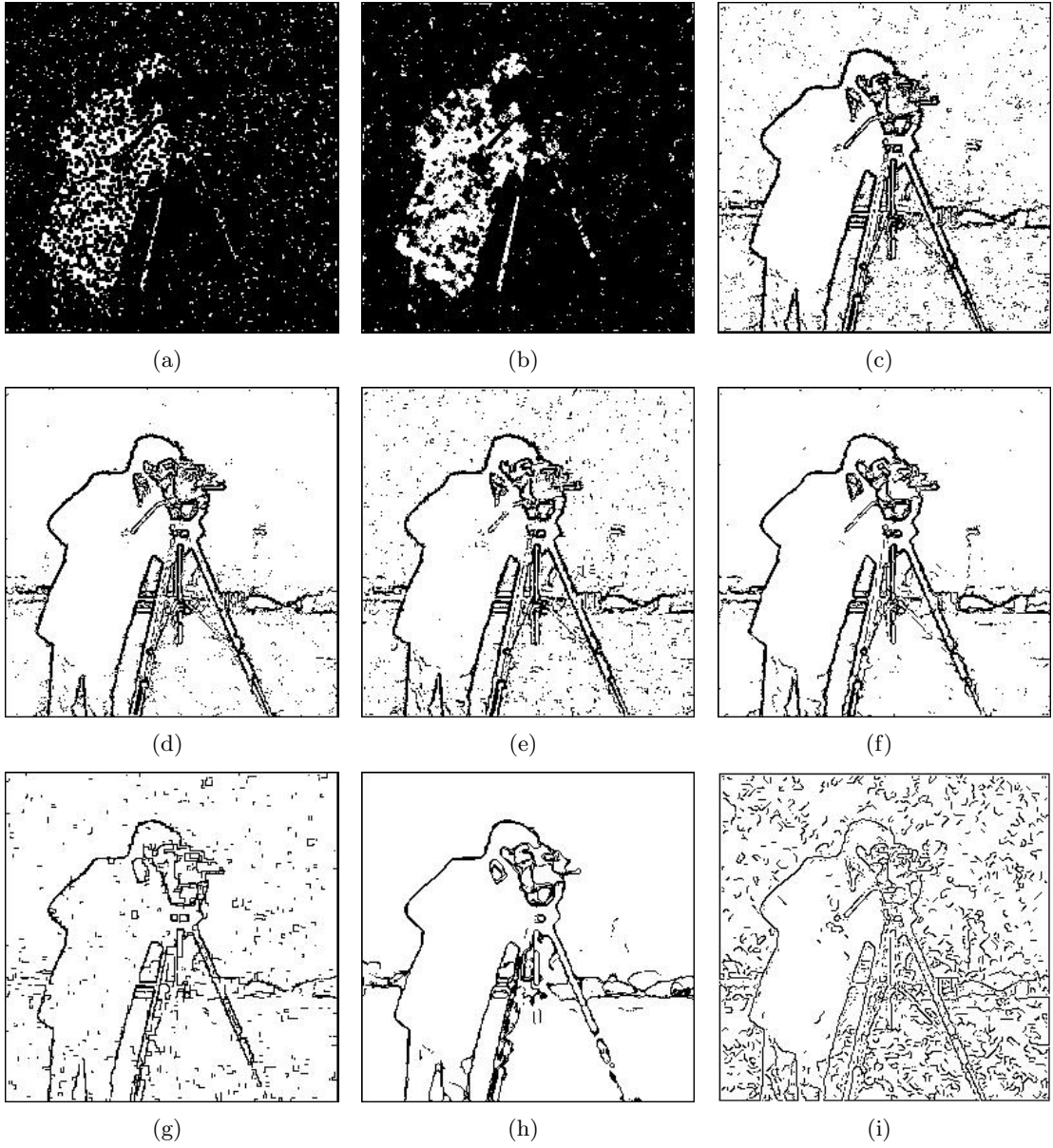

Figure 11: Edge detection result for cameraman image in Figure 8(e): (a) MG detector, (b) Amoeba MG detector, (c) BM detector, (d) Amoeba BM Detector, (e) ATM detector, (f) Amoeba ATM detector, (g) RNM Detector, (h) Amoeba RNM detector, and (i) Canny detector

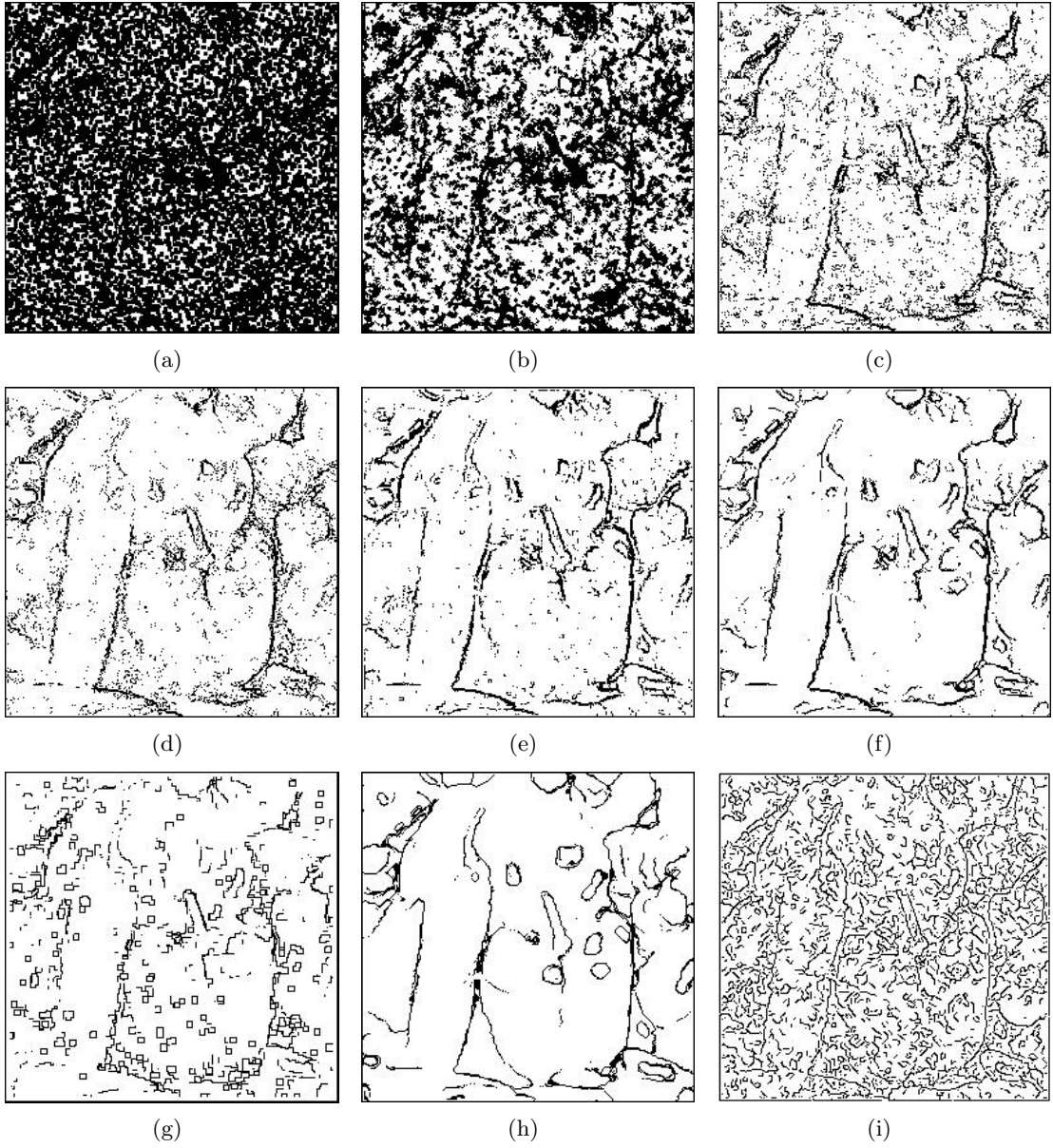

Figure 12: Edge detection result for peppers image in Figure 8(f): (a) MG detector, (b) Amoeba MG detector, (c) BM detector, (d) Amoeba BM Detector, (e) ATM detector, (f) Amoeba ATM detector, (g) RNM Detector, (h) Amoeba RNM detector, and (i) Canny detector